\documentclass[11pt,a4paper]{article}
\usepackage[hyperref]{acl2020}
\usepackage{times}
\usepackage{latexsym}

\usepackage{booktabs}
\usepackage{footnote}
\makesavenoteenv{tabular}
\makesavenoteenv{table}

\usepackage{microtype}

\usepackage{graphicx}
\graphicspath{{images/}}

\usepackage{amsmath}
\aclfinalcopy 

\title{SocialNLP EmotionGIF 2020 Challenge Overview: \\Predicting Reaction GIF Categories on Social Media}

\author{Boaz Shmueli$^{1,2,3,}$\thanks{~~Corresponding author: \href{mailto:shmueli@iis.sinica.edu.tw?Subject=Reactive Supervision: A New Method for Collecting Sarcasm Data}{shmueli@iis.sinica.edu.tw}}~,~ Lun-Wei Ku$^2$ \and Soumya Ray$^3$\\
$^1$Social Networks and Human-Centered Computing, Taiwan International Graduate Program\\
$^2$Institute of Information Science, Academia Sinica\\
$^3$Institute of Service Science, National Tsing Hua University\\
}

\date{}
\begin{document}

\maketitle
\begin{abstract}
We  present  an  overview  of  the  EmotionGIF  2020 Challenge,
held at the 8th International Workshop on Natural Language Processing for Social Media (SocialNLP), in conjunction with ACL 2020.
The challenge  required  predicting  affective reactions to online texts, and 
includes the EmotionGIF dataset, with tweets labeled for the reaction categories.
The novel dataset included 40K tweets with their reaction GIFs. 
Due to the special circumstances of year 2020, two rounds of the competition were
conducted. A total of 84 teams registered for the  task. Of these, 25 teams successfully submitted entries to the evaluation phase in the first round, while 13 teams participated successfully in the second round.  Of the top participants, five teams presented a technical report and shared their code. The top score of the winning team using the Recall@K metric was 62.47\%.

\end{abstract}

\section{Introduction}
Emotions, moods, and other affective states are an essential part of the human experience. The detection
of affective states in texts is an increasingly important area of research in NLP, with important applications
in dialogue systems, psychology, marketing, and other fields \citep{yadollahi2017current}.
Recent approaches have taken advantage of progress in  machine learning, and specifically
deep neural networks \citep{lecun2015deep}, for building models that classify sentiments and emotions in text. Training these models  often requires large amounts of high-quality labeled data. 

Two main
approaches have been used for collection and labeling emotion data: manual annotation and
distance supervision.
With \textit{manual annotation}, humans are presented with a text, and are requested to annotate the text.
When using this approach, several emotional models can be used
for labeling. The two most common models are the discrete emotional model \cite{ekman1971constants}, where the user
needs to select among a few categorical emotions (e.g., disgust, joy, fear), and the dimensional emotion model \cite{mehrabian1996pleasure}, which uses three numerical
dimensions (valence, arousal, dominance) to represent all emotions.

With the help
of crowd-sourcing platforms such as Amazon Mechanical Turk \cite{buhrmester2016amazon}, human annotation can be quickly
scaled up to produce large datasets. However, to achieve large, high-quality datasets,
the cost incurred is usually high. In addition, misinterpretations of text 
due to cultural differences or contextual gaps are common, resulting in unreliable, low-quality labels.
It should be noted that the annotators
detect the \textit{perceived} emotions, i.e., the 
emotions that are recognized in the text.

\begin{figure}
\centering
\fbox{\includegraphics[width=0.75\linewidth]{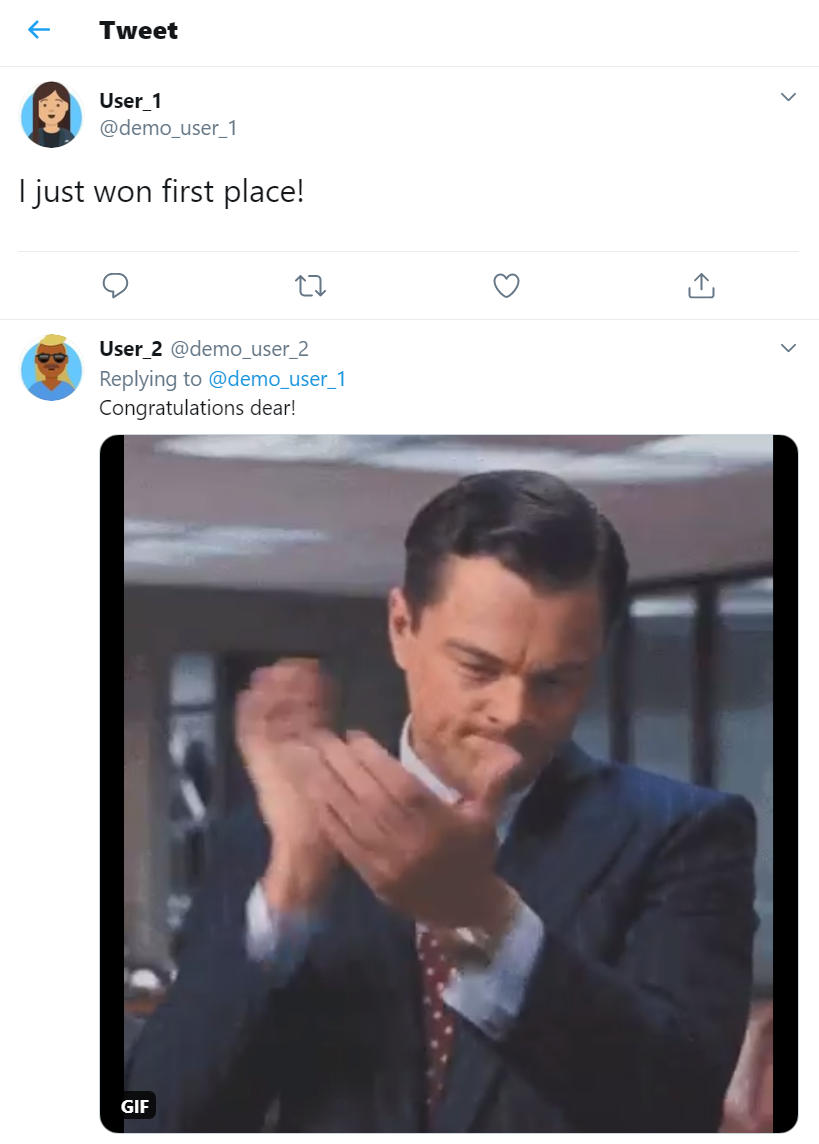}}
\caption{A typical user interaction on Twitter}
\label{figure:leo}
\end{figure}

Another method for data collection is \textit{distant supervision}, often using
emojis
or hashtags (e.g., \citet{go2009twitter}, \citet{mohammad2015using})
This method provides high-volume, automatic
collection of data, albeit with some limitations such as noisy labels  (hashtags might not be related
to the emotions conveyed in the text). It should be noted that the data collected in this case corresponds to the \textit{intended} emotions by the text's author.
Distant supervision can also be used to label 
reactions to text. For example \citet{pool-nissim-2016-distant} use the Facebook feature
that allows users to respond with one of six emojis (Like, Love, Haha,
Wow, Sad and Angry) to collect
posts and their readers' reactions. These reactions are  a proxy to the readers' \textit{induced} emotions -- the emotions they felt when reading the text.
This method is limited by the narrow emotional range of labeling.

To improve research on fine-grained emotional reaction and open up new research possibilities, we conducted the
ReactionGIF 2020 shared task.
The challenge offered a new dataset of 40K tweets with their fine-grained reaction category (or categories).
The task
challenge was to predict each tweet's reactions in an unlabeled evaluation dataset.
In the following sections, we describe and discuss the dataset, the competition, and the results.

\section{EmotionGIF Dataset}
Twitter is a popular micro-blogging site, where users create short text posts known as tweets. In most languages, including English, tweets are limited to 280 characters (the limit is 140 characters in Japanese, Chinese, and Korean). As part of the post, users can also mention other users (@user), and use hashtags (\#hashtag). Additionally, posts can include images videos, or animated GIFs. Animated GIFs are short animations that are commonly used on the internet.
One of the most popular uses of animated GIFs is as reactions in online conversations, 
such as social media interactions.
These GIFs, known
as \textit{reaction GIFs}, are able to convey emotions in an expressive and accurate way \cite{bakhshi2016fast}, and have
become very popular in online conversations. Figure \ref{figure:leo} shows a typical interaction on Twitter:
User\_1 posted a tweet (\textit{"I just won first place!"}),
and User\_2 replied with a tweet that includes
an ``applause''-category reaction GIF, and some text ("Congratulations dear!"). 

\begin{figure}
\centering
\includegraphics[width=0.7\linewidth]{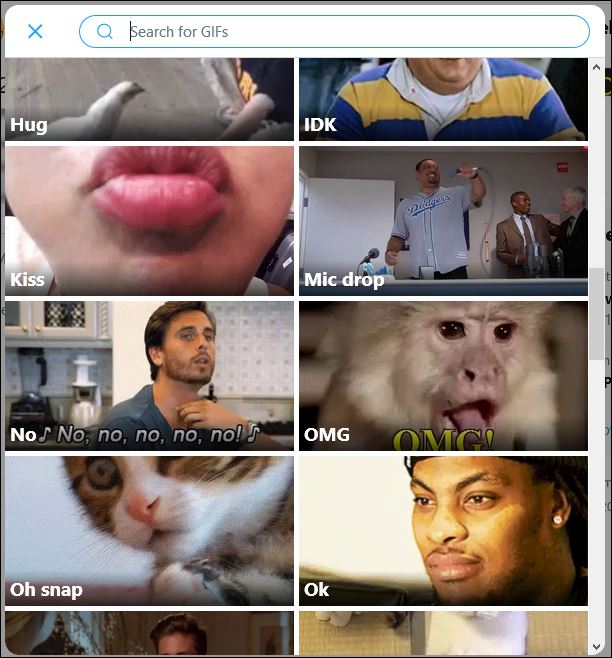}
\caption{A sample of GIF categories on Twitter}
\label{figure:gif-categories}
\end{figure}

\begin{table}
\centering
\resizebox{\columnwidth}{!}{%
\begin{tabular}{@{}llllll@{}}
agree          & happy\_dance & slow\_clap & thank\_you     \\
applause       & hearts       & oops       & thumbs\_down   \\
awww           & high\_five   & please     & thumbs\_up     \\
dance          & hug          & popcorn    & want           \\
deal\_with\_it & idk\footnote{i don't know}          & scared     & win            \\
do\_not\_want  & kiss         & seriously  & wink           \\
eww            & mic\_drop    & shocked    & yawn           \\
eye\_roll      & no           & shrug      & yes            \\
facepalm       & oh\_snap     & sigh       & yolo\footnote{you only leave once}           \\
fist\_bump     & ok           & smh\footnote{shake my head}        & you\_got\_this \\
good\_luck     & omg\footnote{oh my god}          & sorry      &               
\end{tabular}%
}
\caption{GIF categories}
\label{tab:categories}
\end{table}
For the challenge, we collected similar 2-turn interactions. Each sample in the dataset contains the text of the original tweet, the text of the reply tweet,
and the video file of the reaction GIF. 
The label for each tweet is the reaction category (or categories) of the GIF.
Because some replies only contain a reaction GIF, the reply text is optionally empty.
We use a list of 43 reaction categories, pre-defined by the Twitter platform, and used when a user
needs to insert a GIF into a tweet (see Figure \ref{figure:gif-categories}).
The list is shown in Table \ref{tab:categories}, and covers a wide range of emotions,
including love, empathy, disgust, anger, happiness, disappointment, approval, regret, etc.
There is an overlap between  the reaction categories in terms of the GIFs they contain 
and thus some GIFs can belong to more than one category. Consequently, the label may contain more than one reaction category.
For example, GIFs that are categorized with ``shocked'' might also be categorized in ``omg''.
Table \ref{tab:samples} shows a few samples from the training dataset. Note that replies can be optionally empty. The GIF MP4 files are included in the dataset for completeness, but are not
used in this challenge.
\begin{figure}
\centering
\includegraphics[width=\linewidth]{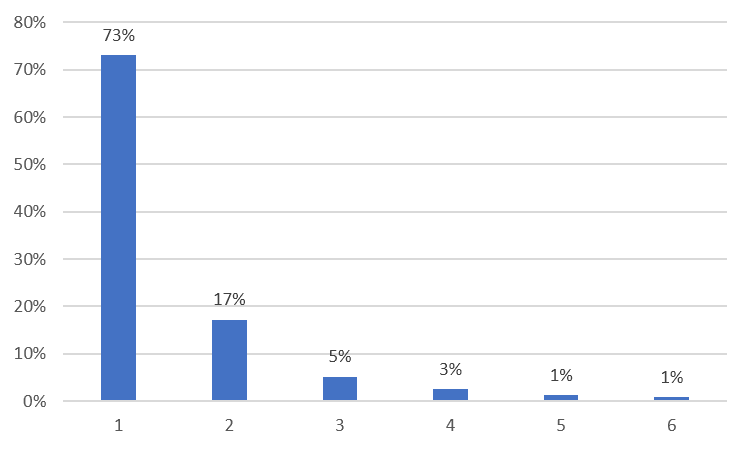}
\caption{Categories per sample}
\label{figure:categories_per_gif}
\end{figure}
\begin{figure*}
\centering
\includegraphics[width=\linewidth]{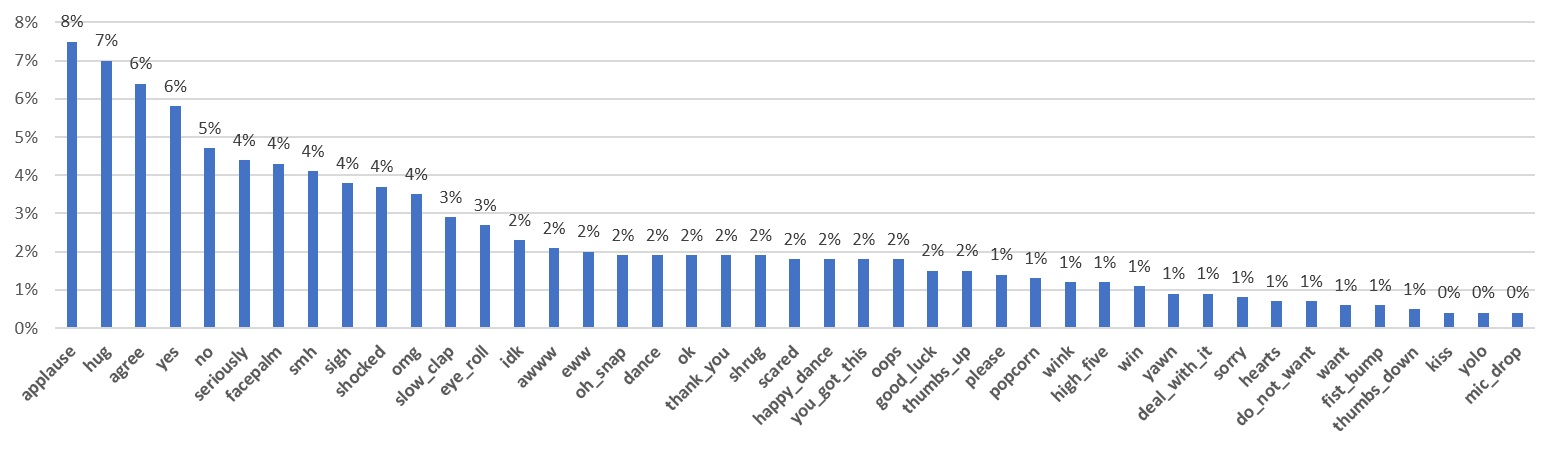}
\caption{User options for Twitter GIF categories}
\label{figure:reaction-distribution}
\end{figure*}
\begin{figure*}
\centering
\includegraphics[width=0.7\linewidth]{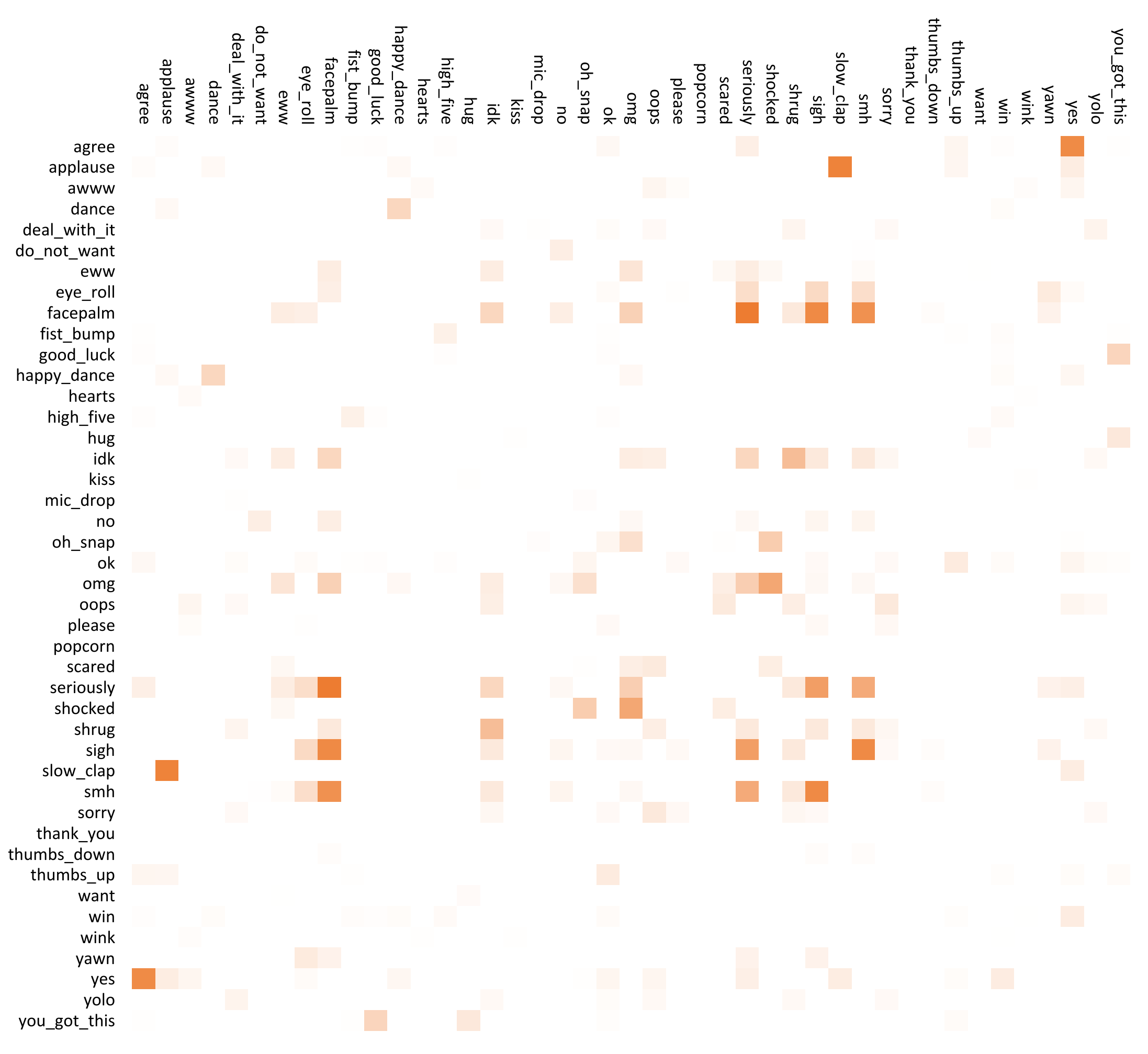}
\caption{GIF category co-occurrence heatmap}
\label{figure:gif-pairs}
\end{figure*}

\begin{table*}
\centering
\resizebox{\textwidth}{!}{%
\begin{tabular}{@{}llll@{}}
\toprule
\textbf{Original tweet} & \textbf{Reply tweet} & \textbf{Reaction GIF} & \textbf{Reaction Categories}\\ \midrule
{Why don’t you interact with me ?} & & d74d...a34e.mp4& oops \\ 
{someone give me a hug (from 2 metres away)} &  &  2be1....5fc0.mp4& want, hug \\
So disappointed in the DaBaby & You’re one to talk & bb2d...cfcf.mp4& smh \\
Bonus stream tonight anyone? & & e91e....49af.mp4& win, ok, thumbs\_up \\
Camila Cabello and You & Of course? & a9fc....b139a.mp4 & shrug, oops, idk \\
\bottomrule
\end{tabular}%
}
\caption{Dataset samples}
\label{tab:samples}
\end{table*}

We collected the EmotionGIF dataset during April 2020, and it includes 40,000 English-language tweets and their GIF reactions. The dataset is divided 80\%/10\%/10\% into \textit{training} (32,000 samples), \textit{development} (4,000 samples), and \textit{evaluation} (4,000 samples) datasets.

\paragraph{Categories per sample} 
Figure \ref{figure:categories_per_gif} shows the distribution of the number of categories per sample.
The majority of samples (73.1\%) in the training dataset
are labeled with a single category. An additional 17.7\% of samples have two labels, and 5.1\% have three categories. The remaining  samples are labeled with 3 to 6 labels, 
\paragraph{Category Distribution}
Figure \ref{figure:reaction-distribution} shows the category distribution. The
categories suffer from uneven distribution; a few of the categories (``applause'', ``hug'', 
``agree'', ``yes'', ``no'') label between 5\% to 10\% of the samples, while most of them label 2\% or
less of the samples.

\paragraph{Category Co-occurrence} The categories are semantically overlapping, and thus some category pairs co-occur more often than others. Figure \ref{figure:gif-pairs} shows
 the co-occurrence heat map. For example, GIFs that are labelled with ``facepalm'' tend to also be labeled with ``seriously'', ``sign'', and ``smh'' (\textbf{S}hake \textbf{M}y \textbf{H}ead), as  these four categories are all expressions related to disappointment. ``shocked'' and ``omg'' (\textbf{O}h \textbf{M}y \textbf{G}od) co-occur frequently, both indicating surprise, etc.

\section{Shared Task}
Due to year 2020's extraordinary circumstances, the competition had two rounds: Round One and Round Two. 
A shared task website was set up\footnote{https://sites.google.com/view/emotiongif-2020/}, which included general information, dates, file format, frequently-asked questions, 
registration form, etc. 
In addition,
two competition websites (Round One, Round Two) were set
up on the Codalab platform\footnote{{https://competitions.codalab.org/}}, where participants could download
the datasets and upload their submissions. 

For the shared task, we provided the training dataset (32K each) with labels, 
and two datasets (development and evaluation, 4K samples each), without labels.
Additionally, the development and evaluation datasets did not contain the video files.
The task was to predict six labels for every sample, with the metric being 
Mean Recall at 6, or $M\!R@6$. To compute $M\!R@6$, we first define the per-sample
recall at 6 for sample $i$, $R@6_i$, which is the ratio:
\begin{equation*}
    R@6_i = \frac{|G_i \cap P_i|}{|G_i|},
\end{equation*}
where $G_i$ is the set of true (``gold'') reaction categories for sample $i$,
and $P_i$ is the set of six predicted reaction categories for sample $i$.
$R@6_i$ is the fraction of reaction categories correctly predicted for sample $i$. 
Because each sample is labeled with a maximum of six categories, $R@6$ is always a value between 0 and 1. 
We then  average over all samples to arrive at $M\!R@6$:

\begin{equation*}
    R=M\!R@6 = \frac{1}{N}\sum_{i=1}^{N} R@6_i
\end{equation*}
We also calculated $R_1$ and $R_2$, which are the Recall at 6 values for the samples with a non-empty reply (i.e., the reply tweet included
both a GIF and text), and with an empty reply (the reply tweet only include a GIF).

Each round of the competition had two phases: practice, and evaluation.
During the practice phase, participants uploaded predictions
for the development dataset and were able to instantly check the performance. 
In the evaluation phase, which determined the competition 
winners, participants uploaded predictions for the evaluation datasets.
Results were hidden until the end of the Round to prevent
overfitting to the data.

\section{Submissions}

\begin{table*}
\centering
\resizebox{\textwidth}{!}{%

\begin{tabular}{lllccc}
\toprule
Rank & Team &  Approach & $R$ & $R_1$ & $R_2$\\ \midrule
1 & Mojitok & Ensemble of transformers (large), label embedding, mixconnect, PAL-N& 62.47 & 61.35 & 63.21\\ 
2 & Whisky &  RoBERTa & 57.31 &  55.22 & 58.70 \\ 
3 & Yankee &Preprocessing, ensemble of transformers (base) & 56.62 & 52.82 & 59.15 \\
4 & Crius & Statistical features, similarity features, BERT, LightGBM & 53.94 & 50.05 & 56.54 \\
5 & IITP & Ensemble of RNNs (CNN, BiLSTM, GRU) & 53.80 & 50.06 & 56.29 \\ \bottomrule
\end{tabular}
}
\caption{Recall at 6 scores for EmotionGIF}
\label{tab:results}

\end{table*}

A total of 84 people registered for
the shared task. 25 teams successfully submitted entries to the evaluation phase in Round One, while 13 teams participated
successfully in Round Two. 
Of the top participants, five teams presented a technical report and shared their code, as was required by the competition rules. 

We provided a simple \textbf{majority baseline} that predicts the 6 most common labels for all samples (applause, hug, agree, yes, no, seriously). $M\!R@6$ for
the majority baseline is 40.0\%.

The highlights of these submissions are summarized below. More details are available in the relevant reports.

\paragraph{Team Mojitok \cite{mojitok}}
This top submission used a combination of methods to attack the challenging aspects of the tasks. 
Four models were used: three  transformer-based models, RoBERTa \cite{liu2019roberta}, DialoGPT \cite{zhang2019dialogpt} and XLNet
\cite{yang2019xlnet}. The fourth was  a RoBERTa model in combination with a label embedding using CGN \cite{chen2019multi} that captures category dependency was also employed. This model were fine-tuned,  using the ``large'' variant of the pretrained models. 5-fold cross validation was used within each model
to produce a total of
20 estimators. Soft-voting ensembles of these estimators were tested in various
combinations. In addition, mixconnect \cite{lee2019mixout} was used for regularization in some of the models. Reduction of the  multi-label problem  Pick-All-Labels Normalised (PAL-N) \cite{menon2019multilabel} multi-class formulation was found to optimize  the $R\!@6$ metric. 
\paragraph{Team Whisky \cite{whisky}}
RoBERTa and  BERT models were evaluated using different hyperparameters,
and with different handling of emojis. The superiority of large RoBERTa model
with long sequences was demonstrated.  
\paragraph{Team Yankee \cite{yankee}}
Elaborate preprocessing was used to increase token coverage. RoBERTa and two BERT models (case and uncased) were fine-tuned and then ensembled using power-weighted sum. The pretrained models are of the ``base'' variant.
Binary cross entropy followed a sigmoid activation layer for prediction. 
\paragraph{Team Crius \cite{crius}}
This method obtained features  by fine-tuned pairwise and pointwise BERT, along with statistical semantic features and similarity-related features. These were
concatenated and  fed into a LightGBM classifier \cite{ke2017lightgbm}.  

\paragraph{Team IITP \cite{iitp}}
Preprocessing included removal of Twitter mentions and replacing emoticons with words. An ensemble of 5 models was  used. The first model used two 2D CNN with attention networks, one for the original tweet and one of the reply tweet, using pre-trained GloVe embeddings. The outputs of the two CNN networks were 
concatenated and fed into a fully-connected layer followed by sigmoid
activation layer and binary cross-entropy. Dropout layers were used at various stages for regularization.
Four additional models with similar top architecture were trained using two instances each of 1D CNN+BiLSTM, Stacked BiGRU, BiLSTM, BiGRU. The outputs from these five models were majority-voted to produce the predictions.
\section{Evaluation \& Discussion}
A summary of the submissions and their challenge scores is available in Table \ref{tab:results}. Presented are the teams that submitted detailed
technical reports and their code for verification. A full leaderboard that includes all the teams is available on the shared task website. 
This section highlights some observations related to the challenge.

\paragraph{Unbalanced Labels.} Emotion detection in text often suffers from a data imbalance problem. Similarly
our dataset (which supervised reactions, not emotions) has a similar phenomenon. 
This would be emphasized if we used a metric that is sensitive to
class imbalances (e.g., Macro-F1). Our metric is less sensitive to these kind of problem. None of the teams decided to take any measures in that regard.

\paragraph{Label dependency}
Multi-label datasets (\citep{tsoumakas2007multi}, \citep{zhang2013review}) introduce
challenging classifications tasks. 
As we can see from Figure \ref{figure:gif-pairs},
in our dataset  the categories are highly dependent. \citep{mojitok} used a new approach to represent this correlation. Classification of multilabel
datasets is a developing area that requires further research.

\paragraph{Models}
All of the submissions used deep learning models. Four of the models were  transformer-based architectures, with most using pre-trained transformers (BERT or its variants). 
Some of the submissions enhanced these model in various ways, e.g. using k-fold CV ensembles \cite{mojitok}. The use of ``large''  vs ``base'' models explained some of the performance differential.

\section{Conclusion}
We summarized the results of the EmotionGIF 2020 challenge, which
was part of the SocialNLP Workshop at ACL 2020. 
The challenge presented a new task that entailed the prediction of affective
reactions to text, using the categories of reaction GIFs as proxies to affective states.
The data included tweets and their GIF reactions. 
We provided brief summaries of each of the eligible participants' entries. Most submissions used transformer-based architectures (BERT, RoBERTa, XL-Net) etc, reflecting their
increasing use in NLP classification tasks, due to their superior performance but also the
availability of easy-to-use programming libraries.
The top system employed the use of various methods, including ensemble, 
regularization, and GCN to to achieve the top score.
\section*{Acknowledgements}
This research was partially supported by the Ministry of Science and Technology of Taiwan under contracts MOST 108-2221-E-001-012-MY3 and MOST 108-2321-B-009-006-MY2.

\bibliographystyle{acl_natbib}
\bibliography{library} 

\begin{thebibliography}{23}
\expandafter\ifx\csname natexlab\endcsname\relax\def\natexlab#1{#1}\fi

\bibitem[{Bakhshi et~al.(2016)Bakhshi, Shamma, Kennedy, Song, De~Juan, and
  Kaye}]{bakhshi2016fast}
Saeideh Bakhshi, David~A Shamma, Lyndon Kennedy, Yale Song, Paloma De~Juan, and
  Joseph~'Jofish' Kaye. 2016.
\newblock {Fast, cheap, and good: Why animated GIFs engage us}.
\newblock In \emph{Proceedings of the 2016 CHI conference on human factors in
  computing systems}, pages 575--586.

\bibitem[{Bi et~al.(2020)Bi, Wang, and Fan}]{crius}
Ye~Bi, Shuo Wang, and Zhongrui Fan. 2020.
\newblock {EmotionGIF-CRIUS}: A hybrid {BERT} and {LightGBM} based model for
  predicting emotion {GIF} categories on {Twitter}.
\newblock Technical report, Ping An Technology (Shenzhen).

\bibitem[{Buhrmester et~al.(2016)Buhrmester, Kwang, and
  Gosling}]{buhrmester2016amazon}
Michael Buhrmester, Tracy Kwang, and Samuel~D Gosling. 2016.
\newblock Amazon's mechanical turk: A new source of inexpensive, yet
  high-quality data?
\newblock \emph{Perspectives on Psychological Science}.

\bibitem[{Chen et~al.(2019)Chen, Wei, Wang, and Guo}]{chen2019multi}
Zhao-Min Chen, Xiu-Shen Wei, Peng Wang, and Yanwen Guo. 2019.
\newblock Multi-label image recognition with graph convolutional networks.
\newblock In \emph{Proceedings of the IEEE Conference on Computer Vision and
  Pattern Recognition}, pages 5177--5186.

\bibitem[{Ekman and Friesen(1971)}]{ekman1971constants}
Paul Ekman and Wallace~V Friesen. 1971.
\newblock Constants across cultures in the face and emotion.
\newblock \emph{Journal of personality and social psychology}, 17(2):124.

\bibitem[{Ghosh et~al.(2020)Ghosh, Roy, Ekbal, and Bhattacharyya}]{iitp}
Soumitra Ghosh, Arkaprava Roy, Asif Ekbal, and Pushpak Bhattacharyya. 2020.
\newblock {EmotionGIF-IITP}: Ensemble-based automated deep neural system for
  predicting category(ies) of a {GIF} response.
\newblock Technical report, IIT Patna.

\bibitem[{Go et~al.(2009)Go, Bhayani, and Huang}]{go2009twitter}
Alec Go, Richa Bhayani, and Lei Huang. 2009.
\newblock Twitter sentiment classification using distant supervision.
\newblock \emph{CS224N project report, Stanford}, 1(12):2009.

\bibitem[{Jeong et~al.(2020)Jeong, Kim, Kim, Moon, and Cho}]{mojitok}
Woojung Choi~Dongseok Jeong, Hyunho Kim, Hyunwoo Kim, Jihwan Moon, and Hyunwoo
  Cho. 2020.
\newblock {EmotionGIF-MOJITOK}: A multi-label classifier with pick-all-models
  ensemble and pick-all-labels normalised loss.
\newblock Technical report, Platfarm Inc.

\bibitem[{Ke et~al.(2017)Ke, Meng, Finley, Wang, Chen, Ma, Ye, and
  Liu}]{ke2017lightgbm}
Guolin Ke, Qi~Meng, Thomas Finley, Taifeng Wang, Wei Chen, Weidong Ma, Qiwei
  Ye, and Tie-Yan Liu. 2017.
\newblock Lightgbm: A highly efficient gradient boosting decision tree.
\newblock In \emph{Advances in neural information processing systems}, pages
  3146--3154.

\bibitem[{LeCun et~al.(2015)LeCun, Bengio, and Hinton}]{lecun2015deep}
Yann LeCun, Yoshua Bengio, and Geoffrey Hinton. 2015.
\newblock Deep learning.
\newblock \emph{nature}, 521(7553):436--444.

\bibitem[{Lee et~al.(2019)Lee, Cho, and Kang}]{lee2019mixout}
Cheolhyoung Lee, Kyunghyun Cho, and Wanmo Kang. 2019.
\newblock Mixout: Effective regularization to finetune large-scale pretrained
  language models.
\newblock In \emph{International Conference on Learning Representations}.

\bibitem[{Liu et~al.(2019)Liu, Ott, Goyal, Du, Joshi, Chen, Levy, Lewis,
  Zettlemoyer, and Stoyanov}]{liu2019roberta}
Yinhan Liu, Myle Ott, Naman Goyal, Jingfei Du, Mandar Joshi, Danqi Chen, Omer
  Levy, Mike Lewis, Luke Zettlemoyer, and Veselin Stoyanov. 2019.
\newblock Roberta: A robustly optimized bert pretraining approach.
\newblock \emph{arXiv preprint arXiv:1907.11692}.

\bibitem[{Mehrabian(1996)}]{mehrabian1996pleasure}
Albert Mehrabian. 1996.
\newblock Pleasure-arousal-dominance: A general framework for describing and
  measuring individual differences in temperament.
\newblock \emph{Current Psychology}, 14(4):261--292.

\bibitem[{Menon et~al.(2019)Menon, Rawat, Reddi, and
  Kumar}]{menon2019multilabel}
Aditya~K Menon, Ankit~Singh Rawat, Sashank Reddi, and Sanjiv Kumar. 2019.
\newblock Multilabel reductions: what is my loss optimising?
\newblock In \emph{Advances in Neural Information Processing Systems}, pages
  10600--10611.

\bibitem[{Mohammad and Kiritchenko(2015)}]{mohammad2015using}
Saif~M Mohammad and Svetlana Kiritchenko. 2015.
\newblock Using hashtags to capture fine emotion categories from tweets.
\newblock \emph{Computational Intelligence}, 31(2):301--326.

\bibitem[{Phen et~al.(2020)Phen, Yang, and Tseng}]{whisky}
Wilbert Phen, Mu-Hua Yang, and Yu-Wun Tseng. 2020.
\newblock {EmotionGIF-Whisky}: {BERT} and {RoBERTa} for emotion classification
  for tweet data.
\newblock Technical report, NCTU.

\bibitem[{Pool and Nissim(2016)}]{pool-nissim-2016-distant}
Chris Pool and Malvina Nissim. 2016.
\newblock \href {https://www.aclweb.org/anthology/W16-4304} {Distant
  supervision for emotion detection using {F}acebook reactions}.
\newblock In \emph{Proceedings of the Workshop on Computational Modeling of
  People{'}s Opinions, Personality, and Emotions in Social Media ({PEOPLES})},
  pages 30--39, Osaka, Japan. The COLING 2016 Organizing Committee.

\bibitem[{Tsoumakas and Katakis(2007)}]{tsoumakas2007multi}
Grigorios Tsoumakas and Ioannis Katakis. 2007.
\newblock Multi-label classification: An overview.
\newblock \emph{International Journal of Data Warehousing and Mining (IJDWM)},
  3(3):1--13.

\bibitem[{Wang et~al.(2020)Wang, Chang, and Tang}]{yankee}
Wei-Yao Wang, Kai-Shiang Chang, and Yu-Chien Tang. 2020.
\newblock {EmotionGIF-Yankee}: A sentiment classifier with robust model based
  ensemble methods.
\newblock Technical report, NCTU.

\bibitem[{Yadollahi et~al.(2017)Yadollahi, Shahraki, and
  Zaiane}]{yadollahi2017current}
Ali Yadollahi, Ameneh~Gholipour Shahraki, and Osmar~R Zaiane. 2017.
\newblock Current state of text sentiment analysis from opinion to emotion
  mining.
\newblock \emph{ACM Computing Surveys (CSUR)}, 50(2):1--33.

\bibitem[{Yang et~al.(2019)Yang, Dai, Yang, Carbonell, Salakhutdinov, and
  Le}]{yang2019xlnet}
Zhilin Yang, Zihang Dai, Yiming Yang, Jaime Carbonell, Russ~R Salakhutdinov,
  and Quoc~V Le. 2019.
\newblock Xlnet: Generalized autoregressive pretraining for language
  understanding.
\newblock In \emph{Advances in neural information processing systems}, pages
  5753--5763.

\bibitem[{Zhang and Zhou(2013)}]{zhang2013review}
Min-Ling Zhang and Zhi-Hua Zhou. 2013.
\newblock A review on multi-label learning algorithms.
\newblock \emph{IEEE transactions on knowledge and data engineering},
  26(8):1819--1837.

\bibitem[{Zhang et~al.(2019)Zhang, Sun, Galley, Chen, Brockett, Gao, Gao, Liu,
  and Dolan}]{zhang2019dialogpt}
Yizhe Zhang, Siqi Sun, Michel Galley, Yen-Chun Chen, Chris Brockett, Xiang Gao,
  Jianfeng Gao, Jingjing Liu, and Bill Dolan. 2019.
\newblock Dialogpt: Large-scale generative pre-training for conversational
  response generation.
\newblock \emph{arXiv preprint arXiv:1911.00536}.

\end{thebibliography}

\end{document}